# Multi-turn Dialog System on Single-turn Data in Medical Domain

Nazib Sorathiya
Computer Science and Engineering
University of California Santa Cruz
Santa Cruz CA USA
nssorath@ucsc.edu

Chuan-An Lin
Computer Science and Engineering
University of California Santa Cruz
Santa Cruz CA USA
clin134@ucsc.edu

Daniel Chen
Computer Science and Engineering
University of California Santa Cruz
Santa Cruz CA USA
dchen61@ucsc.edu

Daniel Xiong
Computer Science and Engineering
University of California Santa Cruz
Santa Cruz CA USA
dxiong@ucsc.edu

Scott Zin
Computer Science and Engineering
University of California Santa Cruz
Santa Cruz CA USA
nzin@ucsc.edu

Yi Zhang
Computer Science and Engineering
University of California Santa Cruz
Santa Cruz CA USA
yiz@soe.ucsc.edu

He Sarina Yang
Department of Pathology and Laboratory Medicine,
Weill Cornell Medicine
New York, New York USA
hey9012@med.coenell.edu

Sharon Xiaolei Huang
Huck Institutes of the Life Sciences,
Penn State University
University Park, PA USA
suh972@psu.edu

## ABSTRACT

Recently there has been a huge interest in dialog systems. This interest has also been developed in the field of the medical domain where researchers are focusing on building a dialog system in the medical domain. This research is focused on the multi-turn dialog system trained on the multi-turn dialog data. It is difficult to gather a huge amount of multi-turn conversational data in the medical domain that is verified by professionals and can be trusted. However, there are several frequently asked questions (FAQs) or single-turn QA pairs that have information that is verified by the experts and can be used to build a multi-turn dialog system.

In this work, we attempt to train a multi-turn conversational dialog system on the single-turn QA pairs. We experiment by using 2 different datasets: QA pair data between patients and doctors from https://www.healthcaremagic.com/ and FAQ data from https://labtestsonline.org/ that is verified and used by the professionals. We further discuss the data, methodology, and results on both datasets in the paper.

## CCS CONCEPTS

- **Applied computing** → Health care information systems

## 1 Introduction

A lot of information is available on the internet which is often not verified and cannot be trusted. This wrong information may have several consequences and especially in the field of medicine, it can be dangerous. Recently there has been a huge interest in developing a chatbot in the medical domain. Much of this research is focused on training these chatbots using multi-turn conversations.

In this work, we focus on creating a multi-turn conversational dialog system that is trained on the single-turn frequently asked questions (FAQs) from a verified source used by experts in the field. We aim to extract information from these FAQs and use it to ask clarification questions to get more information from the user when the user asks an incomplete question.

To further extend our chatbot's ability to utilize more available real-world data, we also conduct experiments on single-turn dialogue data from an online medical consulting platform.

In this scenario, we focus on providing the correct diagnosis given the users' description of their symptoms. Just like what we may encounter during a visit with a doctor, our chatbot will ask clarification questions such as "Do you also have this symptom?" after the patients provide their major concerns in the first turn. By gathering more information from the patients, the chatbot can have a more precise prediction of patients' conditions.

We will use the symptoms and diagnosis appearing in both the patient's description and the doctor's response to formulate a task where we train and evaluate how precisely our system can generate a follow-up clarifying question.

## 2  Related Work

Much research has been done on the multi-turn dialogue system area. [1] focus on the task-oriented dialogue system which requires a knowledge base of the specific task, and [2] deal with a more general scenario. Meanwhile, [3] and [4] focus on conversational question answering where an article is given as the background of the multi-turn question-answering process. All of the above works depend on multi-turn dialogue as training data.

Different perspectives of interactive question answering have also been explored. In the scenario of [5], interaction is achieved by the model that suggests potential follow-up questions after answering the first question. [6] proposes a visual interactive question answering task where the system needs to interact with a real-world-like environment to be able to answer the question. The works of [7] and [8] are most related to our scenario that clarifies ambiguous questions. However, the settings of them rely on either background facts or a knowledge graph.

The paper [9] discusses major three QA approaches, i.e. deep Natural Language Processing (NLP), Information Retrieval (IR) enhanced by shallow NLP, and Template-based QA, and discusses which of these better fit medical applications. The researchers in the paper [10] discuss the use of fuzzy logic for recommendation using social media and social relationships using fuzzy logic. They created fuzzy rules based on the 3 factors: Trust, Similarity, Review for the recommendation. Researchers from the University of Minnesota and Microsoft in the paper [11] Developed a preference elicitation framework to identify which questions to ask a new user to quickly learn their preferences. Their results demonstrate the benefits of starting from offline embeddings and bandit-based explore-exploit strategies.

The system developed by researchers in [12] identifies key factors impacting the recommendation in a healthcare social networking environment and uses semantic web technology and fuzzy logic to represent and evaluate the recommendation. [13] discusses a method for identifying a tailored set of profiles that is acquired by analyzing the implicitly shown preferences of the users. The paper [14] presents an approach that focuses its attention on the group's social interaction during the formulation, discussion, and negotiation of the features the item to be jointly selected should possess. Users are allowed to provide feedback on other's preferences and change significance and bring up new recommendations. The paper [15] proposes a Knowledge-based Recommender Dialog System. With the help of knowledge-grounded information about users' preferences, the dialog system can enhance the performance of the recommendation system.

The majority of the research in multi-turn dialog systems is done using multi-turn dialog datasets and learning about the multi-turn interactions. In the interactive QA systems majority of the systems are focused on getting the correct answers given the context (paragraph) and relevant complete questions. There is very little research done in trying to get the answers for the ambiguous questions in the QA pairs, especially for the FAQs.

## 3  Dataset

### 3.1  FAQ

We have extracted the list of frequently asked question and answer pairs on various lab tests from https://labtestsonline.org/ which is a verified source and is used by the professionals in the field. There are a total of 2529 QA pairs in this extracted dataset. The aim is to find the missing information in the user's question and extract it from the QA pair and ask the clarification questions to get the complete information from the user.

There are various scenarios on how these FAQs can be used to extract information and build a multi-turn dialog system. We distributed these QA pairs in 4 different scenarios:

1. Clarification questions can be asked from information in the question.

2. Clarification questions can be asked from the information in the respective answer.
3. Clarification questions to get the missing entity from user's questions

In scenario 2, asking the clarification question from its respective answers is difficult as there is no access to the answer to the question. In cases where the answer can be extracted from the question in order to ask the clarification question to the user, there will not be any need to ask a question as we can directly provide the answer. Instead of asking a clarification question to the user about the missing entity, it can be resolved through the coreference resolution. There has been a lot of research done for the coreference resolution to solve the problem Scenario 3 tries to fix.

In this study, we are focused on scenario 1. The aim is to identify the information that is missing in the user's query in comparison with the original FAQ and ask clarifying questions to get the missing information. For this study, we filtered the FAQ data to get the FAQ pairs that fit into Scenario 1 and annotated the possible incomplete questions and their respective clarification question manually. On filtering the FAQ data in scenario 1, we were left with around 100 FAQ pairs.

## 3.2 Medical Dialogue

We use the single-turn dialogue data from HealthcareMagic to train and evaluate our system. We collect 200082 pairs of the patient description and doctor response.

To ensure that the single turn dialogue contains enough information so that the doctor can confidently provide a diagnosis given the patient description, we filter the dataset such that it only includes the dialogue where one single diagnosis is proposed by the doctor. We then have 8,494 dialogues remaining.

Sometimes, the doctor's response may repeat the symptoms mentioned in the patient's description, as shown in Figure 1. It means that these symptoms are critical information for the doctor to infer the patient's condition.

Based on this assumption, we reformulate the single-turn medical dialogue record into a multi-turn clarification task, as shown in Figure 1. We recognize the symptoms repeated by the doctor, then remove this information from the patient's description. We only remove the information of randomly one symptom if there are multiple symptoms repeated by the doctor. Since the critical symptom information is hidden, the doctor may need to ask a question to clarify whether the patient has this symptom to make the same diagnosis. Under this scenario, we can measure how precise our system can ask the right clarification question by judging whether it clarifies the same symptom as in the hidden information.

We use all of these dialogue pairs with a repeat as our evaluation data. There are 786 such dialogue pairs in total. The remaining 7,708 pairs are used to first build a probabilistic model for our chatbot to learn the relationship between symptoms and diseases.

We also collected a symptom-disease knowledge base to identify symptom and diagnosis information from the dialogue. The integrated knowledge base contains 131 symptoms and 31 disease terms.

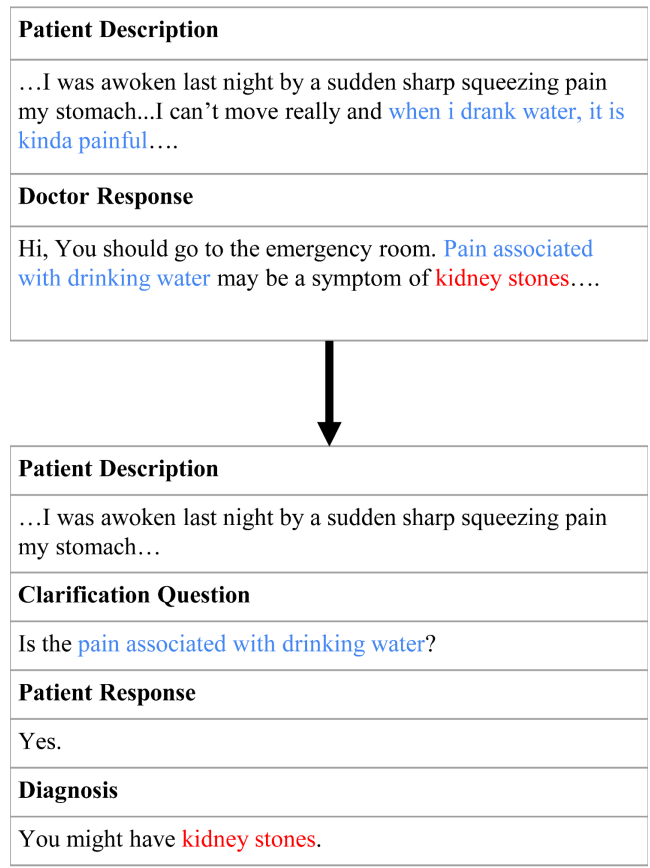

| Patient Description |
|---|
| …I was awoken last night by a sudden sharp squeezing pain my stomach...I can't move really and when i drank water, it is kinda painful…. |

| Doctor Response |
|---|
| Hi, You should go to the emergency room. Pain associated with drinking water may be a symptom of kidney stones…. |

| Patient Description |
|---|
| …I was awoken last night by a sudden sharp squeezing pain my stomach… |

| Clarification Question |
|---|
| Is the pain associated with drinking water? |

| Patient Response |
|---|
| Yes. |

| Diagnosis |
|---|
| You might have kidney stones. |

**Figure 1:** Example from the single-turn medical dialogue dataset (upper part) and the multi-turn dialogue instance (lower part) for evaluation in our clarification task after the conversion process. The text in blue color is the symptom mentioned by both the patient and the doctor. The text in red color is the final diagnosis.

## 4 Methodology

## 4.1 FAQ

From the extracted FAQs, we first identified the FAQs belonging to scenario 1. We then manually annotated the original FAQ pairs into the incomplete user question and its respective clarifying question from the other part of the original FAQ.

Since the dataset size was too small, we decided to start with the rule-based model to first tokenize the original question into 2 parts and then ask the clarifying question from the other part of the original question that was missing from the user's incomplete question.

There are 3 major parts to the rule-based model:

1. **A retrieval program for Matching:** The role of the retrieval program is to match the user's incomplete question with its corresponding original complete question from the dataset.
2. **Tokenization of the extracted original question:** Once the corresponding original complete question is extracted, we tokenize this original question based on the types of the question. Several questions have 2 parts where 1 is the condition mentioned by the user and the other is the question based on the condition. There are also a few questions where the difference between 2 or more entities is asked. For such types of questions, all the entities are extracted and clarification question is generated based on the entities.
3. **Generating clarification question:** Clarification question is generated using the condition part which is missing from the user's incomplete question.

### 4.2 Medical Dialogue

To identify the symptoms and disease information from the dialogue dataset, we use exact matching to extract the identical entries in the knowledge base.

We use the Naive Bayes model to learn the relationship between symptoms and diseases. Given the joint frequency of each diagnosis $d$ and symptom $s$ in the training data, we can estimate the possibility of each diagnosis conditional on the symptoms mentioned in the patient's description based on the assumption of Naive Bayes:

$$P(d|S^p) \propto P(d) \cdot \prod_{s \in S} P(s = I(s \in S^p) | d) \quad (1)$$

Where $S^p$ is the set of symptoms mentioned by the patient $p$, $S$ is the set of all symptoms, $I(s \in S^p) = 1$ if the patient mentioned this symptom, $I(s \in S^p) = 0$ otherwise. We use the Laplacian smoothing method to avoid zero possibility in the probabilistic model parameters.

To generate a clarification question, we want to find the symptom not yet mentioned by the patient but is critical information. We estimate the criticalness of a symptom $s$ by measuring the ratio of the conditional possibilities of the top two diagnoses:

$$\hat{s} = argmax_{s_i \in S, s_i \notin S^p} \frac{P(d_1|S_i^p)}{P(d_2|S_i^p)} \quad (2)$$

where $S_i^p$ is the set of symptoms mentioned by the patient $p$ plus $s_i$, $d_1$ is the diagnosis with the highest probability given $S_i^p$, and $d_2$ is the diagnosis with the second-highest probability.

We expect the $s^\wedge$ we found to match the hidden symptom mentions in our evaluation dataset. To get a better understanding of our system performance, we also evaluate the top 10 $s^\wedge$ rather than just the topmost $s^\wedge$ to observe the trend of recall and precision.

## 5 Results and Discussion

### 5.1 FAQ

We noticed that many of the FAQs for the scenario that we are interested in had a pattern on how they were structured and contained the required information. We implemented a rule-based model to generate the clarification question for different types of questions.

There were also a few questions that did not follow any pattern and contain statements that are very generic and do not have any specific structure. We wanted to use generic rules and not overfit the rules on the questions in the dataset and hence we did not create rules that can be very specific to a single question. With this constraint, we were able to achieve a coverage of 62% on asking clarification questions back to the user as a response to the user's incomplete questions.

### 5.2 Medical Dialogue

Figure 2 shows the performance of our model's symptom clarification prediction from recall@1 to recall@2. For the topmost prediction, our model only achieves 0.0326, which is quite low, and the recall gradually increases to 0.2369 when we consider the top 10 predictions. The low recall@1 score may be due to various reasons. Although there is only one standard target symptom to be clarified based on the original dialogue between the patient and the doctor, there are other possible diseases that the patient could have if we don't know the real condition in advance, and different potential diseases may have different symptoms to clarify. Also, compared to the ideal scenario where it requires multiple turns of clarifying to have a final diagnosis conclusion, we only focus on a single clarification question, making it difficult to hit the target at once. Finally, there is also a data noise issue. We assume that the symptoms repeated by the doctor are critical to the diagnosis, but it is also possible that the doctor mentions the symptom just to give suggestions about how to release this symptom.

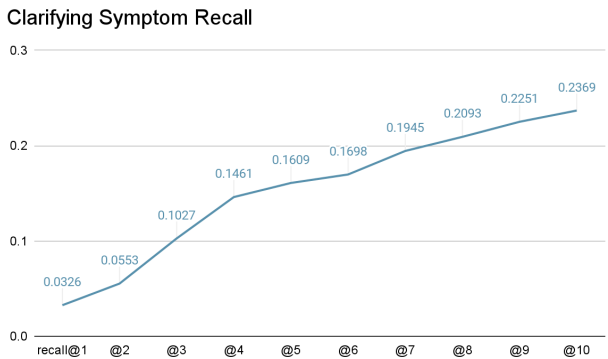

**Figure 2:** Performance of the symptom prediction for clarification from recall@1 to recall@10.

## 6 Challenges and Future Work

In this work, we propose two scenarios to explore the possibility of building a multi-turn dialogue chatbot in the medical domain. We use the FAQ dataset and medical dialogue dataset respectively to build a system that can ask clarification questions to collaborate with the users. We conduct experiments to get preliminary results about how accurately we can generate proper clarification questions, showing the effectiveness of achieving multi-turn dialogue agents based on single turn question-answer pairs.

Compared to the curated FAQ question-answer pairs, the medical dialogue dataset is available on a much larger scale based on its nature of collaborative work by many patients and doctors online. Meanwhile, however, the naturally generated dialogue may contain mixed intentions and redundant information, making it a quite noisy dataset for the purpose of recognizing symptoms and disease relations. Our current approach is severely affected by this noisiness.

In the future, we plan to use the deep learning model to generate clarification after the matching phase. The idea is that the deep learning model will be able to extract the clarification question directly from the original complete question given the user's incomplete question. This will eliminate the 2 step method of tokenization and generation. The challenge with using the deep learning method is to train the model on the small data that we have. Even fine-tuning the pre-trained models will not give significantly good results on the small dataset.

Currently, the model on the FAQ dataset can handle the 2-turn conversation. We will work on Enhancing the model to handle more turns in the conversation.

## ACKNOWLEDGMENTS

This work is supported by the IRKM lab at the University of California, Santa Cruz.